\documentclass[a4paper,fleqn]{cas-sc}

\usepackage{algorithm}
\usepackage{algpseudocode}
\usepackage[numbers]{natbib}
\usepackage{float}

\def\tsc#1{\csdef{#1}{\textsc{\lowercase{#1}}\xspace}}
\tsc{WGM}
\tsc{QE}

\begin{document}
\let\WriteBookmarks\relax
\def\floatpagepagefraction{1}
\def\textpagefraction{.001}
\shortauthors{Md. Sajeebul Islam Sk. et~al.}

\shorttitle{}  


\title [mode = title]{An Explainable Vision-Language Model Framework with Adaptive PID-Tversky Loss for Lumbar Spinal Stenosis Diagnosis}  



%

\author[1]{Md. Sajeebul Islam Sk.} [orcid=0009-0002-3753-6063]

\ead{mdsajeebulislamsk@gmail.com}

\credit{Conceptualization, Methodology, Formal analysis, Investigation, Writing - Original Draft, Visualization}

\author[2,3]{Md. Mehedi Hasan Shawon}[orcid=0000-0002-3883-660X]
\cormark[1]
\ead{mshawon@umd.edu}
\credit{Writing - Review \& Editing, Validation, Supervision, Project administration, Funding acquisition}

\author[1]{Md. Golam Rabiul Alam, PhD}[orcid=0000-0002-9054-7557]
\ead{rabiul.alam@bracu.ac.bd}
\credit{Supervision, Project administration, Funding acquisition}

\affiliation[1]{organization={Department of Computer Science and Engineering},
            addressline={}, 
            city={Dhaka},
            postcode={1212}, 
            state={},
            country={Bangladesh.}}

\affiliation[2]{organization={Department of Computer Science and Engineering},
            addressline={}, 
            city={Dhaka},
            postcode={1212}, 
            state={},
            country={Bangladesh.}}

\affiliation[3]{organization={Department of Electrical and Computer Engineering},
            addressline={}, 
            city={College Park},
            postcode={20742}, 
            state={Maryland},
            country={USA.}}

\cortext[1]{Corresponding author}

\fntext[1]{This project is funded by the Research Seed Grant Initiative-2024, BRAC University. This project is also supported by the Biomedical Science and Engineering Research Center (BIOSE).}


\begin{abstract}
Lumbar Spinal Stenosis (LSS) diagnosis remains a critical clinical challenge, with diagnosis heavily dependent on labor-intensive manual interpretation of multi-view Magnetic Resonance Imaging (MRI), leading to substantial inter-observer variability and diagnostic delays. Existing vision-language models simultaneously fail to address the extreme class imbalance prevalent in clinical segmentation datasets while preserving spatial accuracy, primarily due to global pooling mechanisms that discard crucial anatomical hierarchies. We present an end-to-end Explainable Vision-Language Model framework designed to overcome these limitations, achieved through two principal objectives. We propose a Spatial Patch Cross-Attention module that enables precise, text-directed localization of spinal anomalies with spatial precision. A novel Adaptive PID-Tversky Loss function by integrating control theory principles dynamically further modifies training penalties to specifically address difficult, under-segmented minority instances. By incorporating foundational VLMs alongside an Automated Radiology Report Generation module, our framework demonstrates considerable performance: a diagnostic classification accuracy of 90.69\%, a macro-averaged Dice score of 0.9512 for segmentation, and a CIDEr score of 92.80\%. Furthermore, the framework shows explainability by converting complex segmentation predictions into radiologist-style clinical reports, thereby establishing a new benchmark for transparent, interpretable AI in clinical medical imaging that keeps essential human supervision while enhancing diagnostic capabilities. 
\end{abstract}


\begin{highlights}
\item We propose an end-to-end multimodal Vision-Language framework that integrates advanced models (LLaVA-Med, BiomedCLIP, SmolVLM) to accurately classify the severity of lumbar spinal stenosis, perform precise segmentation, and generate automatic radiology reports directly from MRI scans with clinical text that effectively connects visual pathology with clinical interpretation.
\item We introduce an Adaptive PID-Tversky loss function that efficiently addresses severe class imbalance in medical image segmentation datasets. The model gets better at extracting features from complicated and challenging stenosis patterns by down-weighting readily categorized majority cases with a dynamic, feedback-controlled process.
\item We develop a custom Spatial Patch Cross-Attention module beyond traditional global-pooling methods that discard crucial spatial hierarchies. The method directly extracts spatial patch embeddings from the Vision Transformer (ViT). It then combines these embeddings with textual clinical prompts, allowing for highly accurate, text-guided localization of spinal anomalies.
\end{highlights}

\begin{keywords}
 \sep Vision-Language models \sep Medical image analysis \sep Lumbar Spinal Stenosis Diagnosis \sep Explainable AI
\end{keywords}

\maketitle

\section{Introduction}\label{}
Lumbar spinal stenosis (LSS), a common degenerative musculoskeletal condition, significantly occurs with persistent lower back and leg pain, thereby exerting a substantial burden on global health and socioeconomic structures \citep{batra2025mscanmultistageframeworklumbar} \citep{jcm13237092}. Magnetic Resonance Imaging (MRI) is essential for accurately diagnosing and assessing the severity of a condition. This is because MRI allows clinicians visualize structural problems like narrowing of the spinal canal, protruding discs, and nerve root compression \citep{jcm13237092}. However, the manual interpretation of these complex, multi-view MRI scans is inherently difficult and subject to considerable inter-observer and intra-observer variability, even among experienced radiologists \citep{11211995}\citep{jcm13237092}. As the population ages and the need for MRI diagnoses increases, there's a growing clinical need for diagnostic support studys that are automated, accurate, and cost-effective \citep{batra2025mscanmultistageframeworklumbar}.

In recent years deep learning (DL) and Convolutional Neural Networks (CNNs) have shown significant progress in medical image analysis, including the automated segmentation and classification of spinal structures \citep{11211995} \citep{TANG2022109992}.
 Although conventional single-modal visual networks demonstrate considerable accuracy, their integration into standard clinical workflows suffers from several significant constraints. Initially, conventional models function as "black boxes" \citep{degiorgio2025balancing}, offering only raw classification scores or pixel-wise masks, thereby lacking the transparent rationale essential for medical decision-making \citep{YILIHAMU20252071}\citep{342}. Furthermore, medical datasets are intrinsically characterized by substantial class imbalance \citep{10091194}. Standard loss functions, such as Cross-Entropy, treat all examples uniformly, which leads to models overfitting the majority class and, consequently, inadequately segmenting and classifying intricate, difficult minority cases \citep{9442330} \citep{BUTTNER2024105063}. Conventional models suffer from a singularity of visual features because they rely exclusively on pixel data, they lack the capacity to integrate high-level clinical domain knowledge \citep{10376277}. The inability to connect basic visual patterns with complex diagnostic thinking significantly limits their ability to be used in different clinical situations. This results in a diagnostic performance gap compared to healthcare professionals \citep{jussupow2021augmenting} \citep{degiorgio2025balancing}.

To overcome these research gap, recent studies have increasingly focused on Explainable AI (XAI) and Multimodal Large Language Models (MLLMs) \citep{YILIHAMU20252071} \citep{342}. Vision-Language Models (VLMs) use visual data alongside natural language to connect complex visual features with a single semantic space that includes textual medical information. This allows the network to analyze scans using built-in clinical criteria \citep{nam2025multimodal}. Moreover, VLMs are capable of Automated Radiology Report Generation (ARRG), thereby converting these multimodal features into coherent, easily understandable clinical narratives \citep{jcm13237092} \citep{342}. While frameworks exploring generative image-to-text approaches for the lumbar spine have shown promise, they frequently struggle with precise diagnostic extraction and the nuanced semantic alignment required for varying degrees of stenosis \citep{tumko2024neural} \citep{galbusera2018generative}. Furthermore, none of the existing end-to-end models simultaneously address the fundamental challenge of feature extraction under extreme class imbalance while maintaining linguistic explainability.

To address these challenges, we propose a novel end-to-end multimodal Vision-Language framework for the highly accurate, explainable diagnosis of LSS. We conduct a comprehensive benchmark of cutting-edge VLMs (LLaVA-Med \citep{llavamed}, BiomedCLIP \citep{biomedclip}, and SmolVLM \citep{smolvlm}) utilizing spatial patch embeddings and cross-attention feature fusion. To address the critical segmentation bottleneck caused by imbalanced pathological data \citep{10.1145}, we formulate a novel Adaptive PID-Tversky Loss function. Inspired by proportional-integral-derivative (PID) \citep{PID} control theory with Tversky index \citep{tverskylossfunction}, this dynamic loss formulation autonomously adjusts its penalty weights based on global false-positive and false-negative rates during training, forcing the network to focus its learning capacity on the hardest examples (moderate and severe stenosis). The precision-segmented features are subsequently routed through a specialized ARRG module to generate multi-line, radiologist-style clinical reports, effectively bridging the gap between quantitative AI prediction and qualitative clinical explanation. In summary, our contributions are as follows:

\begin{itemize}
    \item  We propose an end-to-end multimodal Vision-Language framework that integrates advanced models (LLaVA-Med, BiomedCLIP, SmolVLM) to accurately classify the severity of lumbar spinal stenosis, perform precise segmentation, and generate automatic radiology reports directly from MRI scans with clinical report that effectively connects visual pathology with clinical interpretation.

    \item We introduce an Adaptive PID-Tversky loss function that efficiently addresses severe class imbalance in medical image segmentation datasets. The model gets better at extracting features from complicated and challenging stenosis patterns by down-weighting readily categorized majority cases with a dynamic, feedback-controlled process.
        
    \item We develop a custom Spatial Patch Cross-Attention module beyond traditional global-pooling methods that discard crucial spatial hierarchies. The method directly extracts spatial patch embeddings from the Vision Transformer (ViT). It then combines these embeddings with textual clinical prompts, allowing for highly accurate, text-guided localization of spinal anomalies.
    
    \item We design an integrated Automated Radiology Report Generation (ARRG) module. This module translates localized segmentation masks, percentages of affected areas, and severity logits into clear, multi-line clinical reports. This significantly enhances the transparency and explainability of the computer-aided diagnosis (CAD) study.
    
    \item We thoroughly validate our framework with different VLM models using a balanced dataset, performing a full clinical and linguistic evaluation in both computer vision metrics (Accuracy, Dice, IoU) and natural language generation metrics (BLEU, ROUGE, METEOR, CIDEr).
\end{itemize}

Section 2 provides a summary of the relevant literature. Section 3 demonstrates a detailed explanation of the proposed materials and methods, including data preparation, the Spatial Patch Cross-Attention VLM architecture, and the mathematical formulation of the Adaptive PID-Tversky loss. The following section 4 describes the experimental setup. Section 5 presents the experimental results obtained using the proposed framework. Section 6 provides a discussion of the results, compares them to current research, and acknowledges these limitations. The final conclusions of this study are presented in Section 7.

\section{ Related work}\label{}

Early attempts to automate the diagnosis of lumbar spinal stenosis (LSS) were largely dependent on conventional machine learning techniques, which were integrated with manual feature engineering. Koh and colleagues developed a diagnostic technique utilizing magnetic resonance myelography (MRM). This method incorporated binarization and edge detection for dural sac segmentation, succeeded by a two-level classifier. The diagnostic accuracy achieved by this approach was 91.3\% \citep{Mukku2025ArtificialIL}. Similarly, Ruiz-España and his team utilized signal intensity segmentation and B-spline curve fitting to quantitatively evaluate dural sac diameter ratios, yielding a sensitivity of 70\% and a specificity of 81.7\% \citep{Mukku2025ArtificialIL}\citep{TANG2022109992}. A notable constraint of these conventional techniques is their reliance on manual pre-segmentation and the extraction of manually defined features. As a result, these methods are time-consuming, susceptible to inter-observer variability, and difficult to apply across varied multi-institutional clinical datasets.

To overcome the limitations of manual feature engineering and reduce reliance on expensive MRI equipment, Suzuki et al. \citep{SUZUKI20242086} developed a convolutional neural network (CNN) algorithm to identify lumbar spinal stenosis (LSS) requiring surgery, using standard lumbar radiographs. Plain X-rays couldn't show soft tissues as well as MRI scans \citep{SUZUKI20242086}, they used multitask learning. This allowed them to predict the levels needing surgery and assess the spinal canal area at the same time \citep{SUZUKI20242086}.  used a modified convolutional neural network (CNN) architecture with depthwise convolutions and Grad-CAM visualization successfully identify important features in the intervertebral joints and posterior discs \citep{SUZUKI20242086}. Consequently, their approach achieved an Area Under the Curve (AUC) of 0.89 and an accuracy of 84\% in external validation, as reported by Mukku et al \citep{Mukku2025ArtificialIL}. Despite this accuracy, the researchers acknowledged a significant limitation in their data. To reduce potential bias, the L5/S1 level was excluded from the analysis. This decision was based on the unusual stenosis patterns often seen in that area \citep{SUZUKI20242086}. Consequently, the model's applicability was constrained by its dependence on subjective surgical criteria derived from a single institution, a limitation the  intended to address in subsequent versions by employing multi-institutional datasets for fine-tuning \citep{SUZUKI20242086}. To mitigate the architectural constraints inherent in analyzing variable MRI slice sequences, Batra et al. introduced M-SCAN, a complex multistage framework that employs multi-view cross-attention for the grading of spinal canal stenosis \citep{batra2025mscanmultistageframeworklumbar}. A notable limitation of prior 3D CNNs was their rigid need for a uniform slice count per patient, which failed to accurately mirror the heterogeneity observed in real-world clinical datasets \citep{batra2025mscanmultistageframeworklumbar}. To mitigate this constraint, Batra and colleagues developed a sequence-based architecture that processes axial and sagittal images independently, employing pre-trained EfficientNet encoders, followed by Bidirectional Gated Recurrent Units (Bi-GRUs) and LSTM layers \citep{batra2025mscanmultistageframeworklumbar}. M-SCAN showed high performance with different numbers of slices, achieving a high AUROC of 0.971 and a 93.80\% accuracy by using a spatial dropout and cross-attention mechanism to combine features \citep{batra2025mscanmultistageframeworklumbar}. In addition, to address the significant class imbalance in their grading dataset, the  used a Weighted Cross-Entropy (WCE) loss function to give more importance to misclassifications of the minority class \citep{batra2025mscanmultistageframeworklumbar}.

In the domain of dense anatomical segmentation, a persistent limitation of deep learning models has been the catastrophic loss of fine structural details caused by the arbitrary resizing of high-resolution medical images to fit standard network inputs \citep{Mukku2025ArtificialIL}. To address the problem of resolution loss, Mukku et al. developed an advanced AI pipeline. This pipeline initially used a Super-Resolution Convolutional Neural Network (SRCNN). The SRCNN was used to convert low-resolution patches into a high-dimensional feature space, which was then used for segmentation \citep{Mukku2025ArtificialIL}. Following this, SegNet was employed to segment the spinal canal, and a Convolutional Block Attention Module (CBAM) was incorporated to refine the features, culminating in the application of a Swin Transformer for the ultimate classification task. By exploiting the Swin Transformer's shifted window mechanism, they avoided the limitations of fixed receptive fields that are characteristic of conventional Vision Transformers (ViTs), thus attaining a notable overall accuracy of 95.2\% and an F1-score of 96.12\% \citep{Mukku2025ArtificialIL}. In a similar process, Ghobrial and Roth focused on spatial precision, proposing a deep learning methodology that employed the MultiResUNet architecture for the automated segmentation and quantification of the dural sac cross-sectional area (DSCA) \citep{10.3389/fradi.2025.1503625}. Recognizing the persistent issues of inter-radiologist variability and significant time demands associated with traditional manual DSCA measurement, their automated model generated highly objective geometric outputs. Their MultiResUNet produced improved spatial overlap, featuring a Pearson correlation coefficient of 0.9917 and a remarkably low Mean Absolute Error (MAE) of 23.7 mm² on the primary dataset and an accuracy of 99.9\% during external validation \citep{10.3389/fradi.2025.1503625}. However, the use of T1-weighted axial images represented a  drawback. The range of imaging sequences required for full clinical application was not included in this approach. The  proposed to overcome this in future research by incorporating T2-weighted multi-sequence data alongside domain adaptation techniques to broaden the model's flexibility \citep{10.3389/fradi.2025.1503625}. Lin et al. also addressed the quantification of particular spinal indices through the introduction of an object-specific bi-path network (OSBP-Net) \citep{AlAntari2025MRI}. To mitigate the uncover segmentation boundaries produced by conventional static loss functions, they incorporated a novel cross-space distance-preserving regularization technique into their objective function. This integration improved the precise measurement of the dural sac and intervertebral disc mid-sagittal diameters, resulting in minimal prediction errors and the disc diameter exhibited an error of 1.27 ± 0.08 mm \citep{AlAntari2025MRI}.

Conventional deep learning models are typically black box which limit their interpretability despite their high spatial and category precision. To address the limitations and lack of transparency inherent in vision-only models, recent cutting-edge research has shifted towards Multimodal Vision-Language Models (VLMs) and Large Language Models (LLMs). Yilihamu et al. proposed GPT4LFS framework, a sophisticated multimodal fusion study was developed to improve the classification of lumbar foraminal stenosis \citep{YILIHAMU20252071}. To overcome the limitations of isolated visual feature extraction, their model employed a pre-trained ConvNeXt to process high-dimensional MRI features concurrently with expert medical descriptions generated by GPT-4o and encoded through RoBERTa. By utilizing the Mamba architecture during the feature fusion stage, GPT4LFS effectively coordinated the spatiotemporal relevance between the imaging features and textual semantics. The model showed a diagnostic accuracy of 93.7\%, with a sensitivity of 95.8\%, and a Cohen’s Kappa of 0.89 \citep{YILIHAMU20252071}.However, the researchers acknowledged that the substantial computational and financial demands associated with the GPT-4o foundation model constrained their capacity to perform more extensive cohort analyses \citep{YILIHAMU20252071}. In addition, the researchers recognized the constraint inherent in relying only on sagittal MRI slices. Author proposed that subsequent iterations would improve this by integrating axial and coronal perspectives, thereby advancing a more comprehensive representation of the pathological data \citep{YILIHAMU20252071}.

Dong et al. building on the potential of clinical text integration, presented a BERT-based spinal LLM designed to improve the accuracy of classifying intricate lumbar disorders, such as degenerative disc disease and spondylolisthesis \citep{dong2024classification}. A fundamental constraint of conventional CNN classifiers lies in their failure to contextualize localized pixel data within the framework of broader, patient-specific biomechanical parameters. To overcome this drawback, Dong et al. utilized early fusion layers to unite CNN-extracted MRI texture features with explicitly tokenized numerical spinal measurements (such as the lumbar lordotic angle and specific disc heights) \citep{dong2024classification}. By passing these integrated tokens through BERT's multi-head self-attention mechanism, the network dynamically modeled the interdependencies between anatomical variables, allowing the study to classify 61 distinct combinations of lumbar disorders with accuracy, precision, and recall metrics consistently nearing 0.90 \citep{dong2024classification}.While highly robust, the authors recognized a critical limitation: the model's heavy reliance on expert-confirmed MRI cases for ground-truth benchmarking introduced potential human diagnostic bias \citep{dong2024classification}. To mitigate this bias and the risk of domain overfitting, they successfully conducted external validation on an independent dataset of 514 cases, and they proposed the future development of automated, standardized data preprocessing pipelines to handle varying MRI acquisition protocols across international, multi-center environments \citep{dong2024classification}.

\section{Methodology} \label{}

\begin{figure}[p]
    \centering
    \includegraphics[
        width=\textwidth,
        height=0.90\textheight,
        keepaspectratio
    ]{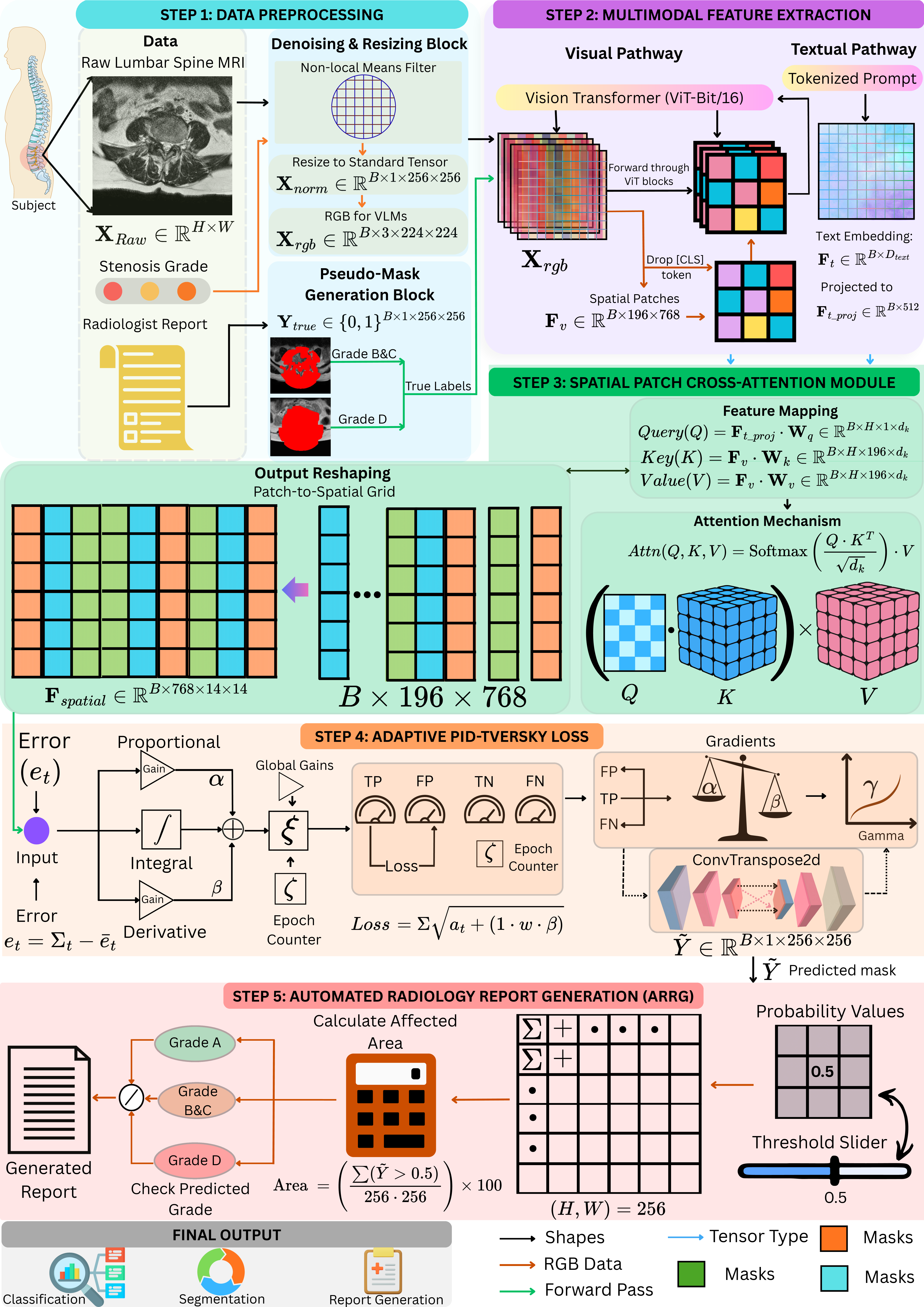}
    \caption{Detailed Model Architecture, the proposed multimodal vision-language framework for Lumbar Spinal Stenosis (LSS) diagnosis.}
    \label{fig:workflow}
\end{figure}

In this section, we explain the architecture and computational steps of our proposed Vision-Language model framework for diagnosing Lumbar Spinal Stenosis (LSS). A detailed workflow is shown in Figure \ref{fig:workflow}, the entire pipeline is divided into five main stages along with dataset definition: (1) data preprocessing, augmentation, and pseudo-mask generation (2) multimodal feature extraction (3) spatial patch cross-attention, (4) Adaptive PID-Tversky loss function and (5) automated radiology report generation.

\subsection{Dataset}

The LSS MRI Dataset \citep{LSS_dataset} is a publicly accessible benchmark dataset for the analysis of lumbar spine MRI data. The dataset consists of anonymized clinical MRI data for patients who are suffering from back pain. There are a total of 515 patients in the dataset. The dataset covers the lowest three vertebrae, the lowest three intervertebral discs, and the sacral area. There are a total of 48,345 images in the dataset. The resolution of the images in the dataset is 320 × 320 pixels with 12-bit pixel precision.


\clearpage
\restoregeometry

\subsection{Data Preprocessing, Augmentation, and Pseudo-Mask Generation} \label{}

As shown in Figure \ref{fig:workflow}, the initial raw MRI input image is represented as a three-channel tensor \(\mathbf{X}_{\mathrm{raw}} \in \mathbb{R}^{3 \times H \times W}\), where \(H\) and \(W\) represent the height and width of the spatial resolution, respectively. Magnetic resonance imaging (MRI) images suffer from scanner-induced artifacts and background noise that can obscure fine anatomical structures \citep{islam2023improving}, we apply a Non-Local Means (NLM) \citep{manjon2009adaptive} denoising filter to the raw images. Unlike traditional local filters that simply blur adjacent pixels together, our NLM algorithm restores a pixel by computing a weighted average of all pixels in the image, where the weights are determined by the structural similarity of their local neighborhood patches. Mathematically, we formulate the filtered value at pixel \(i\), denoted by \(\mathrm{NLM}(i)\) \citep{manjon2009adaptive}:
\begin{equation}
\mathrm{NLM}(i)=\frac{1}{Z(i)}\sum_{j\in I} w(i,j)\,x_{\mathrm{raw}}(j)
\label{eq:nlm_main}
\end{equation}
where \(x_{\mathrm{raw}}(j)\) denotes the intensity of pixel \(j\), \(Z(i)=\sum_{j} w(i,j)\) is the normalizing constant, and the weight \(w(i,j)\) is defined as \citep{manjon2009adaptive}:
\begin{equation}
w(i,j)=\exp\left(-\frac{\|N_i-N_j\|_2^2}{h^2}\right)
\label{eq:nlm_weight}
\end{equation}
where \(h\) is the filtering parameter that controls the decay rate. In lumbar spine MRI, we find NLM filter effectively removes high-frequency Gaussian and Rician noise while preserving the sharp, high-contrast edges of the intervertebral discs, thecal sac, and nerve roots required for accurate stenosis grading. As the LSS dataset \citep{LSS_dataset} contain imbalanced class data, we implement data augmentation approach on the training dataset. By implementing control geometric and photometric alterations such as random rotations within a \(\pm 3^\circ\) range, \(5\%\) zoom scaling, horizontal flipping, and contrast modifications (\(\pm 10\%\)) \citep{bagui2021resampling}, we increase the minority classes and balance the dataset.

The lack of comprehensive pixel-level annotations created by human experts is one of the main challenges in medical image segmentation \citep{li2024lvit}. Due to the unavailability of manually segmented images for our dataset, we utilize the clinical radiologist reports linked to each MRI scan to guide both our selected Vision-Language Model (VLM) and pseudo-ground truth generation. The detailed workflow summarized in Algorithm~\ref{alg:pseudo_mask}, adaptive pseudo-mask generation is performed by combining Otsu's thresholding \citep{Otsu} and Canny edge detection \citep{canny} to identify anatomical boundaries around the spinal canal region. The subsequent morphological refinement is conditioned on the severity described in the radiologist's report: for severe stenosis (Grade~D), stronger dilation with \(9 \times 9\) kernels followed by closing is applied to capture the compressed canal area, whereas for mild to moderate cases (Grades B\&C), smaller dilation kernels and localized erosion are used to avoid overestimation \citep{wu2016improved}. In this way, the text-based radiologist reports directly impact mask construction, enabling to generate reliable pseudo-masks and integrate expert clinical knowledge into the automated segmentation framework.

\begin{algorithm}[t]
\caption{Adaptive pseudo-mask generation}
\label{alg:pseudo_mask}
\begin{algorithmic}[1]
\State \textbf{Input:} $\mathbf{X}_{\mathrm{raw}}, r$ \Comment{$\mathbf{X}_{\mathrm{raw}}$: raw MRI image, $r$: radiologist report}
\State \textbf{Output:} $\mathbf{M}$ \Comment{$\mathbf{M}$: pseudo-mask}
\State $\mathbf{X}_g \gets \mathrm{Gray}(\mathbf{X}_{\mathrm{raw}})$ \Comment{$\mathbf{X}_g$: grayscale image}
\State $\mathbf{B} \gets \mathrm{Otsu}(\mathbf{X}_g)$ \Comment{$\mathbf{B}$: threshold}
\State $\mathbf{E} \gets \mathrm{Canny}(\mathbf{X}_g)$ \Comment{$\mathbf{E}$: edge}
\State $\mathbf{R} \gets \mathbf{B} \cup \mathbf{E}$ \Comment{$\mathbf{R}$: initial region}
\State $s \gets \mathrm{Severity}(r)$ \Comment{$s$: stenosis severity from report}
\If{$s=\mathrm{D}$}
    \State $\mathbf{R} \gets \mathrm{Close}(\mathrm{Dilate}(\mathbf{R},\,9\times9))$ \Comment{severe stenosis refinement}
\Else
    \State $\mathbf{R} \gets \mathrm{Erode}(\mathrm{Dilate}(\mathbf{R},\,k_s))$ \Comment{$k_s$: small kernel for mild/moderate cases}
\EndIf
\State $\mathbf{M} \gets \mathrm{Mask}(\mathbf{R})$ \Comment{final pseudo-mask}
\end{algorithmic}
\end{algorithm}

\subsection{Multimodal Feature Extraction} \label{}

We use advanced Vision-Language Models (VLMs) such as BiomedCLIP, SmolVLM, and LLaVA-Med to overcome the limitations of conventional Convolutional Neural Networks (CNNs) \citep{cnnlrp}, which rely only on visual pixels. These models inherently function within a joint vision-language semantic space. As medical image segmentation requires an in-depth knowledge of both anatomical structures and clinical details \citep{li2024lvit}, we use this multimodal architecture to extract features through two concurrent channels: a text channel for clinical knowledge and a vision channel for spatial features.

For the text channel, we define clinical text prompts corresponding to different grades of stenosis severity, capturing expert radiologist descriptions ranging from normal anatomical structures to severe thecal sac compression. The prompts are tokenized and processed using specialized text encoders: PubMedBERT for BiomedCLIP \citep{biomedclip}, SmolLM2 for SmolVLM \citep{smolvlm}, and the Vicuna-based LLM for the LLaVA-Med \citep{llavamed} implementation, resulting in a dense semantic text embedding $F_t \in \mathbb{R}^{B \times D_{\text{text}}}$. To resolve the inherent dimensional differences across encoders, we pass these embeddings through a linear projection layer, mapping them into a standardized shared dimensional space $D_{\text{shared}}$. Simultaneously, the three-channel MRI tensors \(X_{rgb} \in \mathbb{R}^{B \times 3 \times 224 \times 224}\) are processed through the visual channel utilizing a Vision Transformer (ViT) backbone. For LLaVA-Med, the CLIP-ViT-L/14 encoder \citep{ViT-L/14} is employed. To address the typical Vision Language Model (VLM) dependency on the global classification token ([CLS]), which compromises the fine-grained local spatial details crucial for delineating spinal stenosis boundaries \citep{liu2024vision}, we modify the extraction process to exclude the [CLS] token during the forward pass. Instead, we extract the spatial patch embeddings \citep{liu2024vision}. For input processed into a 14×14 patch grid, this extraction results in precisely 196 distinct spatial patch tokens, denoted as \(F_v \in \mathbb{R}^{B \times 196 \times D_{vis}}\). By retaining these localized patch features, we capture the spatial features of the spinal canal for establishing the essential foundation for the Cross-Attention module.

\subsection{Spatial Patch Cross-Attention Module}

We design a custom 8-head Spatial Patch Cross-Attention module to ensure that the medical text effectively facilitates the visual segmentation process. This module addresses the semantic gap by directing the network's focus towards specific physical regions of the MRI slice that precisely correspond to the provided clinical descriptions. We utilize the dense semantic text embeddings $F_t$ to generate the Queries $(Q)$, while the 196 spatial patch embeddings $F_v$ serve as the Keys $(K)$ and Values $(V)$. Subsequently, we apply linear projections to map the text and image features into a shared attention dimension $d_k$:

\begin{equation}
Q = F_t W_q,\qquad K = F_v W_k,\qquad V = F_v W_v
\end{equation}

where $W_q$, $W_k$, and $W_v$ are learnable weight matrices. Then, the cross-attention mechanism is computed using the standard scaled dot-product formulation \citep{sk2025}:

\begin{equation}
\mathrm{Attention}(Q, K, V) = \mathrm{Softmax}\left(\frac{QK^T}{\sqrt{d_k}}\right)V
\end{equation}

In this process, the model computes attention scores to evaluate the relevance of each spatial patch in relation to the clinical text prompt. The text-weighted visual features are then processed through an output projection layer and combined with the original spatial patch embeddings using a residual connection. 

The attended feature sequence $F_{\text{attn}} \in \mathbb{R}^{B \times 196 \times D_{\text{vis}}}$ remains a one-dimensional sequence. As medical segmentation requires dense 2D pixel-level predictions \citep{gu2019cenet}, we reshape this sequence back into a geometric spatial grid based on the original patch layout. By computing 196 patches, we map the tensor into a 2D format, resulting in the text-guided spatial feature map $F_{\text{spatial}} \in \mathbb{R}^{B \times D_{\text{vis}} \times 14 \times 14}$. Finally, to reshape this spatial grid to the original image dimensions, we pass $F_{\text{spatial}}$ through a Convolutional Neural Network (CNN) decoder. This decoder consists of a sequential series of transposed convolutions, batch normalization, and ReLU activations to upsample the feature map \citep{yuan2022dcaunet}. Through this process, the model reconstructs the precise morphological boundaries of the spinal structures denoted as the final predicted segmentation mask $\hat{Y}$.

\subsection{Adaptive PID-Tversky Loss Function}

We develop an adaptive loss function based on a closed-loop Proportional-Integral-Derivative (PID) controller \citep{PID} to dynamically optimize the segmentation objective during training. The controller calculates a correction \(u_t\) from three control components: the proportional response to the current error ($e_t$), the integral of accumulated past errors $(\sum e_i)$, and the derivative representing the rate of error change $\left(\Delta e_t = e_t - e_{t-1}\right)$. The equation is defined as \citep{zhao2017pid}:
\begin{equation}
    u_t = K_p e_t + K_i \sum e_i + K_d \Delta e_t   
\end{equation}

where $K_p$, $K_i$, and $K_d$ are the proportional, integral, and derivative, respectively. At the end of each training epoch $t$, the model calculates the predicted segmentation masks against the ground-truth masks and computes the total number of False Positives (FP) and False Negatives (FN) over the training set. These two errors are used to define a normalized imbalance signal:
\begin{equation}
    e_t = \frac{FN - FP}{FN + FP + \epsilon}   
\end{equation}

where $\epsilon$ a small constant added to avoid numerical instability and $e_t$ is the current segmentation bias of the model. Based on the control output $u_t$, the adaptive class weights are updated for the next epoch as follows \citep{spadon2021pay}:
\begin{equation}
    \alpha_{t+1} = 1 - \beta_{t+1}, \qquad
    \beta_{t+1} = \beta_t + u_t, \qquad
    \beta_{t+1} \in [0.65,\,0.85]
\end{equation}

where $\beta$ is adaptively updated within a bounded range for stable optimization and $\alpha$ is adjusted as the class weight. When the model predicts more False Negatives, the controller increases $\beta$ instead of relying on fixed class weights. Using the updated parameters $\alpha_{t+1}$ and $\beta_{t+1}$, the Tversky Index (TI) is computed as \citep{ti}:
\begin{equation}
    TI = \frac{TP + \epsilon}{TP + \alpha_{t+1}FP + \beta_{t+1}FN + \epsilon}
\end{equation}

where $TP$ denotes the number of True Positives. To enhance learning in challenging regions, we add a focal weight parameter $\gamma$ \citep{abraham2023focal} integrated with the Tversky loss function. The final loss function is \citep{abraham2018novelfocaltverskyloss}:
\begin{equation}
    Loss_{PT} = (1 - TI)^\gamma
\end{equation}

where $Loss_{PT}$ denotes the PID-Tversky loss. The proposed loss function adjusts weights according to the model's segmentation errors, reduces attention on background pixels, and focuses more effectively on challenging stenotic boundary regions.

\subsection{Automated Radiology Report Generation}

We design an Automated Radiology Report Generation (ARRG) module for generating a radiology-style textual report from the input MRI image $x$ and the predicted segmentation mask $\hat{Y}$. First, the visual encoder extracts a high-level feature representation $f_v$ and a dedicated mask encoder processes $\hat{Y}$ into a latent morphological representation $f_m$ that encodes structural deformation patterns associated with spinal canal stenosis. In parallel, the module calculates quantitative measurements from the predicted mask to estimate the extent of stenotic involvement. The affected area percentage is computed as:
\begin{equation}
\text{Affected Area (\%)} = \left( \frac{\sum (\hat{Y} > 0.5)}{H \times W} \right) \times 100
\end{equation}
where $H$ and $W$ denote the spatial dimensions of the target region. The ARRG module then maps the estimated affected area to a severity template: Grade A for affected area $< 10\%$, Grade B\&C for $10\%{-}25\%$, and Grade D for $> 25\%$, \citep{takahashi1989chronic} corresponding to no significant, mild-to-moderate, and severe spinal canal compression, respectively. The aggregated representation $f_r$, the language generation component predicts a report $R = \{w_1, w_2, \dots, w_n\}$ token by token as \citep{huang2023kiutknowledgeinjectedutransformerradiology}:
\begin{equation}
P(R \mid x, \hat{Y}) = \prod_{t=1}^{n} P(w_t \mid w_{<t}, f_r)
\end{equation}
where $w_t$ denotes the $t$-th generated token and $w_{<t}$ represents the previously generated tokens. During training, the report generation module is optimized using an autoregressive negative log-likelihood loss \citep{huang2023kiutknowledgeinjectedutransformerradiology}:
\begin{equation}
\mathcal{L}_{\text{ARRG}} = - \sum_{t=1}^{n} \log P(w_t^{*} \mid w_{<t}^{*}, f_r)
\end{equation}
where $w_t^{*}$ denotes the ground-truth token at step $t$. Finally, by leveraging visual features, mask-derived morphology, and quantitative severity mapping, the ARRG module produces clinically relevant radiology reports.


\section{Results}

Our study focuses on four primary areas to comprehensively assess the clinical effectiveness and reliability of the multimodal vision-language framework we develop. The fields of study include evaluating stenosis severity, precisely segmenting anatomical features at the pixel level, autonomously producing radiology reports, and conducting a comprehensive ablation study to thoroughly assess the framework's performance.

\subsection{Experimental Setup} \label{}

All experiments are performed using an NVIDIA GeForce RTX 4070 GPU with 12 GB VRAM. We use PyTorch 2.5.1+cu121, Hugging Face Transformers, OpenCV, scikit-learn, and PEFT libraries to develop the framework, and Automatic Mixed Precision (AMP) with torch.bfloat16 is enabled to reduce GPU memory usage and accelerate training.

We adopte QLoRA \citep{qlora} for parameter-efficient adaptation, using AdamW \citep{AdamW} optimizer with a batch size of 8, weight decay of 0.01, and a learning rate of $1 \times 10^{-4}$. For segmentation, the encoder and decoder are fine-tuned with learning rates of $1 \times 10^{-5}$ and $1 \times 10^{-3}$, respectively. The model is trained for up to 30 epochs using the proposed Adaptive PID-Tversky Loss. Early stopping is applied if the validation Dice score is unchanged for 5 epochs. For report generation, the model is trained for 50 epochs with an AdamW optimizer, a learning rate of $5 \times 10^{-4}$, a weight decay of $5 \times 10^{-5}$, a gradient clipping of 1.0, and Focal Loss with $\gamma = 2.0$. Every hyperparameter was selected empirically to achieve optimal findings.

\subsection{Evaluation Matrix} \label{}

We evaluate the proposed framework through its three consecutive tasks: pixel-level segmentation, diagnostic classification, and automated report generation. For segmenting spinal structures, we determine the Dice Similarity Coefficient (DSC), Intersection over Union (IoU), and Hausdorff distance \citep{vu}. In the stenosis classification task, we measure Accuracy, Precision, Recall (Sensitivity), Specificity, and F1-score \citep{f-1}. Furthermore, for automated report generation to assess semantic alignment, lexical diversity, and exact textual similarity with the ground truth radiologist reports, we calculate BLEU (1-4), ROUGE-1, ROUGE-2, ROUGE-L, METEOR, CIDEr scores, Jaccard index, TF-IDF Cosine similarity, and Distinct-n (Dist-1, Dist-2) metrics \citep{le2025}.

\subsection{Classification Results}

In the initial stage of our workflow, diagnostic classification is crucial to guide therapeutic interventions through accurate classification of patient data. A comprehensive analysis of the models shows different structural tradeoffs between diagnostic accuracy and computational efficiency. 

\begin{table}[htbp]
\centering
\caption{Diagnostic metrics for Vision-Language Models in the classification of lumbar spinal stenosis. The table shows precision, recall, F1-score and accuracy for three anatomical grades (A: normal, B\&C: mild-to-moderate, D: severe) using three different model types: LLaVA-Med, BiomedCLIP and SmolVLM. Bold values indicate the best-performing model for each metric.}
\label{tab:classification_performance}
\small
\begin{tabular}{llccccc}
\toprule
\textbf{Model} & \textbf{Evaluation Metric} & \textbf{Grade A} & \textbf{Grade B\&C} & \textbf{Grade D} & \textbf{Macro Average} & \textbf{Accuracy} \\ 
\midrule
LLaVA-Med & Precision & \textbf{0.9366} & \textbf{0.8495} & 0.9387 & \textbf{0.9083} & \textbf{90.69\%} \\
          & Recall    & 0.8892 & \textbf{0.8890} & 0.9425 & \textbf{0.9069} & \\
          & F1-Score  & 0.9123 & \textbf{0.8688} & \textbf{0.9406} & \textbf{0.9072} & \\ 
\midrule
BiomedCLIP & Precision & 0.9225 & 0.7867 & \textbf{0.9407} & 0.8833 & 87.77\% \\
           & Recall    & \textbf{0.9124} & 0.8740 & 0.8468 & 0.8777 & \\
           & F1-Score  & \textbf{0.9175} & 0.8280 & 0.8913 & 0.8789 & \\ 
\midrule
SmolVLM    & Precision & 0.8900 & 0.7950 & 0.7960 & 0.8270 & 82.66\% \\
           & Recall    & 0.8820 & 0.6530 & \textbf{0.9440} & 0.8263 & \\
           & F1-Score  & 0.8860 & 0.7170 & 0.8640 & 0.8223 & \\
\bottomrule
\end{tabular}
\end{table}

The LLaVA-Med model shows high diagnostic performance, achieving an accuracy of 90.69\% (95\% Confidence Interval (CI): 88.2\%-92.9\%) and a macro F1-score of 0.9072. This result exceeds the lightweight SmolVLM model ($p < 0.001$, McNemar's test \citep{McNemar}), which achieves an accuracy of 82.66\% (95\% CI: 79.8\%-85.1\%). The precision of LLaVA-Med is correlated to its large parameter count and advanced linguistic framework that offering the computational power necessary to distinguish minor nerve root compressions. In contrast, BiomedCLIP provides a highly effective clinical solution. By utilizing specialized pre-training model, BiomedCLIP effectively aligns visual features with clinical text by achieving an accuracy of 87.77\% (95\% CI: 85.3\%-89.8\%) and reduces computational costs. 

\begin{figure}
    \centering
    \includegraphics[width=0.6\columnwidth]{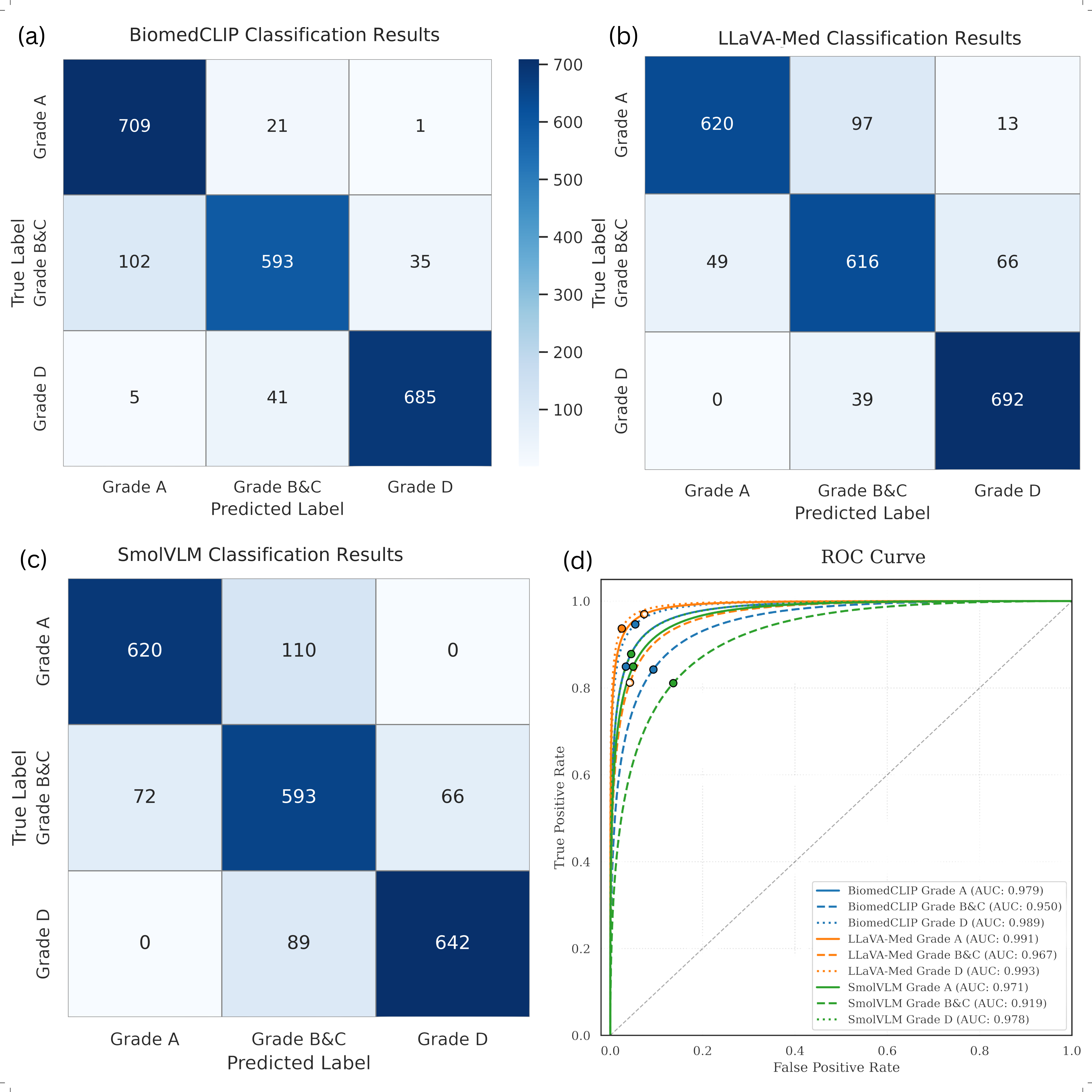}
    \caption{Classification performance comparison across multi-modal VLM models. (a–c) Confusion matrices from the clinical test set for BiomedCLIP, LLaVA-Med, and SmolVLM, respectively, displaying predicted versus true severity grades (A: normal, B\&C: mild-to-moderate stenosis, D: severe stenosis). (d) Receiver operating characteristic (ROC) curves quantifying model discrimination performance across severity grades.}
    \label{fig:confusion_matrices}
\end{figure}

From a clinical viewpoint, it is important to accurately detect patients who might need surgery, while avoiding false positives. The reliability improves sorting and reduces important diagnostic delays. The confusion matrices in Figure \ref{fig:confusion_matrices} show all models accurately identify severe clinical grades with notable reliability. Normal spinal canals (Grade A) include broad, free thecal sacs, whereas severe stenosis (Grade D) appears as a noticeable structural anomaly. As a result, LLaVA-Med effectively identified 685 severe cases and 709 normal cases with high sensitivity. This high discriminative capacity is further corroborated by the Receiver Operating Characteristic (ROC) analysis. LLaVA-Med achieved an exceptional Area Under the Curve (AUC) of 0.993 for severe Grade D cases and 0.991 for normal Grade A cases, underscoring its robustness in isolating definitive surgical candidates and healthy patients. BiomedCLIP and SmolVLM also demonstrated strong predictive power for these severe cases, yielding AUCs of 0.989 and 0.978, respectively.

Mild and moderate grades present considerable diagnostic difficulty. The class variance in mild-to-moderate Grade B\&C stenosis is very high, leading to small boundary changes that mask the differences between healthy and diseased conditions. The problem is clearly visible in the SmolVLM model, which faced difficulties in mapping these complex morphological variations of its limited parameter set. Consequently, it often misclassifies normal and severe cases as mild-to-moderate stenosis 110 and 89 times, respectively. BiomedCLIP faces minor difficulties in this ambiguous area, incorrectly classifying normal instances as mild-to-moderate stenosis on 97 times. These findings suggest that although computational intelligence studies excel in identifying anatomical extremes to enhance workflow, supervision of clinical radiologists is essential for assessing degenerative adjustments.

\subsection{Segmentation Results}

In Table \ref{tab:segmentation_global}, spatial segmentation of stenotic spinal regions evaluate across three vision-language models using pixel-level metrics including Dice Similarity Coefficient (DSC), Intersection over Union (IoU), precision, recall, and specificity. The use of high-resolution spatial patches allows advanced anatomical border detection across all analyzed models. LLaVA-Med shows higher segmentation performance, with the maximum Dice Similarity Coefficient of 0.9624 (95\% CI: 0.951–0.973) and statistical significance compare to SmolVLM (p < 0.05, Wilcoxon signed-rank test \citep{Wilcoxon}). BiomedCLIP get high Dice score 0.9512 (95\% CI: 0.938–0.964), indicating a small performance difference of 0.0112 absolute points relative to LLaVA-Med. This small difference highlights the similar spatial alignment abilities of the two parameter-dense models. Precision measures supported this order, with LLaVA-Med reaching 0.9655 and BiomedCLIP reaching 0.9570, which is a very small difference of 0.0085. In contrast, recall (sensitivity) metrics show better convergence, with LLaVA-Med (0.9610) slightly outperforming BiomedCLIP (0.9574) by only 0.0036. The nearly same sensitivity values suggest that both models have outstanding true positive detection rates for stenotic tissue borders, effectively identifying authentic diseased regions without considerable over-segmentation. However, the lightweight SmolVLM design shows significant performance decline across all global criteria. SmolVLM achieve a Dice score of 0.8603 (95\% CI: 0.841–0.879) with an associated IoU of 0.7566. Looking at LLaVA-Med (the best performer) and SmolVLM (the worst performer), we see a big difference in their Dice scores (0.01021) and their IoU numbers (0.1722). SmolVLM's precision significantly decrease to 0.8402, reflecting a 0.1253 absolute point shortfall compared to LLaVA-Med. This suggests that parameter-efficient models, although adept at identifying general anatomical areas, exhibit low precision in specifying complex pixel-level boundaries when compare with their higher-parameter models.

\begin{table}[htbp]
\centering
\caption{Pixel-level segmentation performance metrics for Vision-Language Models in lumbar spinal stenosis. The table reports Dice Score with 95\% Confidence Intervals (CI), Intersection over Union (IoU), Precision, Recall (Sensitivity), and Specificity across LLaVA-Med, BiomedCLIP, and SmolVLM. Bold values indicate the best-performing model for each metric.}
\label{tab:segmentation_global}
\begin{tabular}{lccccc}
\toprule
\textbf{Model} & \textbf{Dice Score (CI)} & \textbf{IoU} & \textbf{Precision} & \textbf{Recall} & \textbf{Specificity} \\ 
\midrule
LLaVA-Med & \textbf{0.9624 (0.951-0.973)} & \textbf{0.9288} & \textbf{0.9655} & \textbf{0.9610} & \textbf{0.9412} \\
BiomedCLIP & 0.9512 (0.938-0.964) & 0.9147 & 0.9570 & 0.9574 & 0.9279 \\
SmolVLM & 0.8603 (0.841-0.879) & 0.7566 & 0.8402 & 0.8841 & 0.8567 \\
\bottomrule
\end{tabular}
\end{table}

A sequential study of segmentation performance, categorized by clinical severity grade in Table \ref{tab:segmentation_grade_wise}, demonstrate significant behavioral variability and performance decline patterns dependent on severity among the tested models. Macro-averaged global indicators frequently miss significant performance regressions in complex clinical scenarios; therefore, grade-specific analysis is crucial for providing necessary clinical context. In cases of severe surgical impingements (Grade D), spatial accuracy consistently show remarkable precision across the best models. BiomedCLIP achieve a Dice score of 0.9816 and an IoU of 0.9680, slightly above LLaVA-Med's Grade D Dice coefficient of 0.9800 by 0.0016. The similarity in severe case performance indicates that both models have developed strong feature representations for clear disease addresses. In standard anatomical presentations (Grade A), BiomedCLIP shows complete accuracy in localization, attaining a perfect recall (sensitivity) of 1.0000, in contrast to LLaVA-Med's 0.9650. This outstanding results highlights BiomedCLIP's ability for precise anatomical differentiation between normal and pathological tissue, achieving 0\% false negative detections in healthy cases. The shift to extremely ambiguous mild-to-moderate situations (Grade B\&C) revealed significant performance differences among the models. BiomedCLIP suffer severe performance regression, decreasing from a Dice score of 0.9816 in Grade D to 0.8965 in Grade B\&C. This is a significant internal decline of 0.0851 absolute points. This significant decline in performance demonstrates the morphological nuances evident in early-stage stenosis, where the limits of the ligamentum flavum become unclear and anatomical differences are masked. LLaVA-Med achieve a Grade B\&C Dice score of 0.9422, resulting in an internal performance loss of only 0.0378 compared to its Grade D baseline of 0.9800.  Notably, SmolVLM, although having the lowest global performance metrics, shows the strongest cross-grade stability. The model obtain Dice scores of 0.8598 (Grade A), 0.8609 (classes B and C), and 0.8601 (Grade D), demonstrating exceptional consistency with a maximum variance of only 0.0011 throughout all severity classes. This remarkable consistency, although achieve at lower absolute accuracy, indicates that the spatial cross-attention mechanism ensures consistent morphological grounding regardless of disease complexity, offering interpretable spatial stability as an additional benefit to peak performance.

\begin{table}[htbp]
\centering
\caption{Quantitative evaluation of segmentation performance across clinical severity grades (A: normal, B\&C: mild-to-moderate, and D: severe). The table reports Dice Score, Intersection over Union (IoU), and Sensitivity for three Vision-Language Models. Bold values indicate the best-performing model for each metric and grade.}
\label{tab:segmentation_grade_wise}
\begin{tabular}{llccc}
\toprule
\textbf{Model} & \textbf{Metric} & \textbf{Grade A} & \textbf{Grade B\&C} & \textbf{Grade D} \\ 
\midrule
LLaVA-Med & Dice Score & 0.9650 & \textbf{0.9422} & 0.9800 \\
          & IoU & \textbf{0.9310} & \textbf{0.8954} & 0.9600 \\
          & Sensitivity & 0.9650 & \textbf{0.9380} & \textbf{0.9800} \\ 
\midrule
BiomedCLIP & Dice Score & \textbf{0.9754} & 0.8965 & \textbf{0.9816} \\
           & IoU & 0.9056 & 0.8705 & \textbf{0.9680} \\
           & Sensitivity & \textbf{1.0000} & 0.8955 & 0.9765 \\ 
\midrule
SmolVLM    & Dice Score & 0.8598 & 0.8609 & 0.8601 \\
           & IoU & 0.7560 & 0.7577 & 0.7560 \\
           & Sensitivity & 0.8890 & 0.8836 & 0.8796 \\
\bottomrule
\end{tabular}
\end{table}

\begin{figure}
    \centering
    \includegraphics[width=0.8\columnwidth]{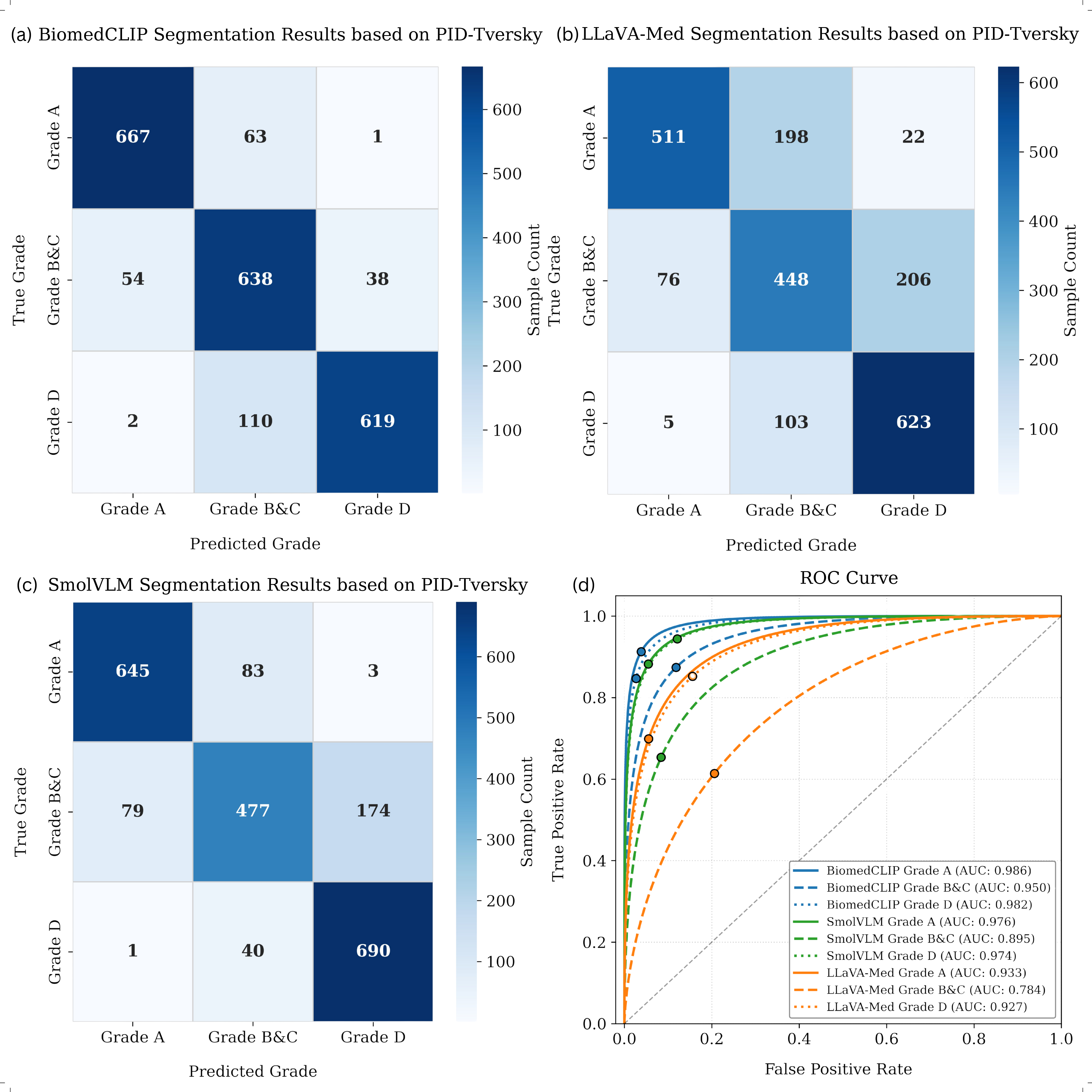}
    \caption{Segmentation-based severity classification performance across multi-modal VLM models. (a–c) Confusion matrices mapping pixel-level segmentation outputs to clinical severity grades for BiomedCLIP, LLaVA-Med, and SmolVLM (all trained with the proposed Adaptive PID-Tversky loss). (d) Receiver operating characteristic (ROC) curves quantifying the models' spatial discrimination performance derived from segmentation masks across clinical severity levels.}
    \label{fig:segmentation_history}
\end{figure}

Segmentation-to-classification mapping, which is shown in Figure \ref{fig:segmentation_history}, allows to make the diagnostic reliability of segmentation-generated spatial boundaries visible. BiomedCLIP shows good discrimination at all severity levels by properly detecting 667 Grade A, 638 Grade B\&C, and 619 Grade D occurrences. The model keeps the boundaries clear, only mistakenly labeling 54 Grade B\&C cases as Grade A and 38 Grade D cases as Grade B\&C. This high level of spatial consistency is supported by outstanding ROC performance, with Area Under the Curve (AUC) values of 0.986 (Grade A), 0.950 (Grade B\&C), and 0.982 (Grade D). LLaVA-Med, on the other hand, has the greatest global Dice coefficient but has difficulty finding intermediate diagnostic areas. The model only accurately classifies 448 Grade B\&C cases, which means it is missing 190 true positives compared to BiomedCLIP. It often gets 198 Grade B\&C cases wrong and thinks they are typical (Grade A) and overestimates severity by putting 206 Grade D instances in the Grade B\&C category. So, the AUC for the diagnostically unclear Grade B\&C class drops to 0.784, which is a substantial difference of 0.166 (26.3\% relative reduction) from BiomedCLIP's AUC of 0.950 for the same class. LLaVA-Med has excellent pixel-level accuracy but lower class-level discrimination shows that it makes anatomically accurate segmentations that rarely line up with clinically defined severity thresholds. SmolVLM provides a decent performance of classifying, with 477 true positives for Grade B\&C and an AUC of 0.895 for this middle class. However, it has a tendency to under-segment severe cases, misclassifying 174 Grade D instances as Grade B\&C.

\begin{figure}
    \centering
    \includegraphics[width=\columnwidth]{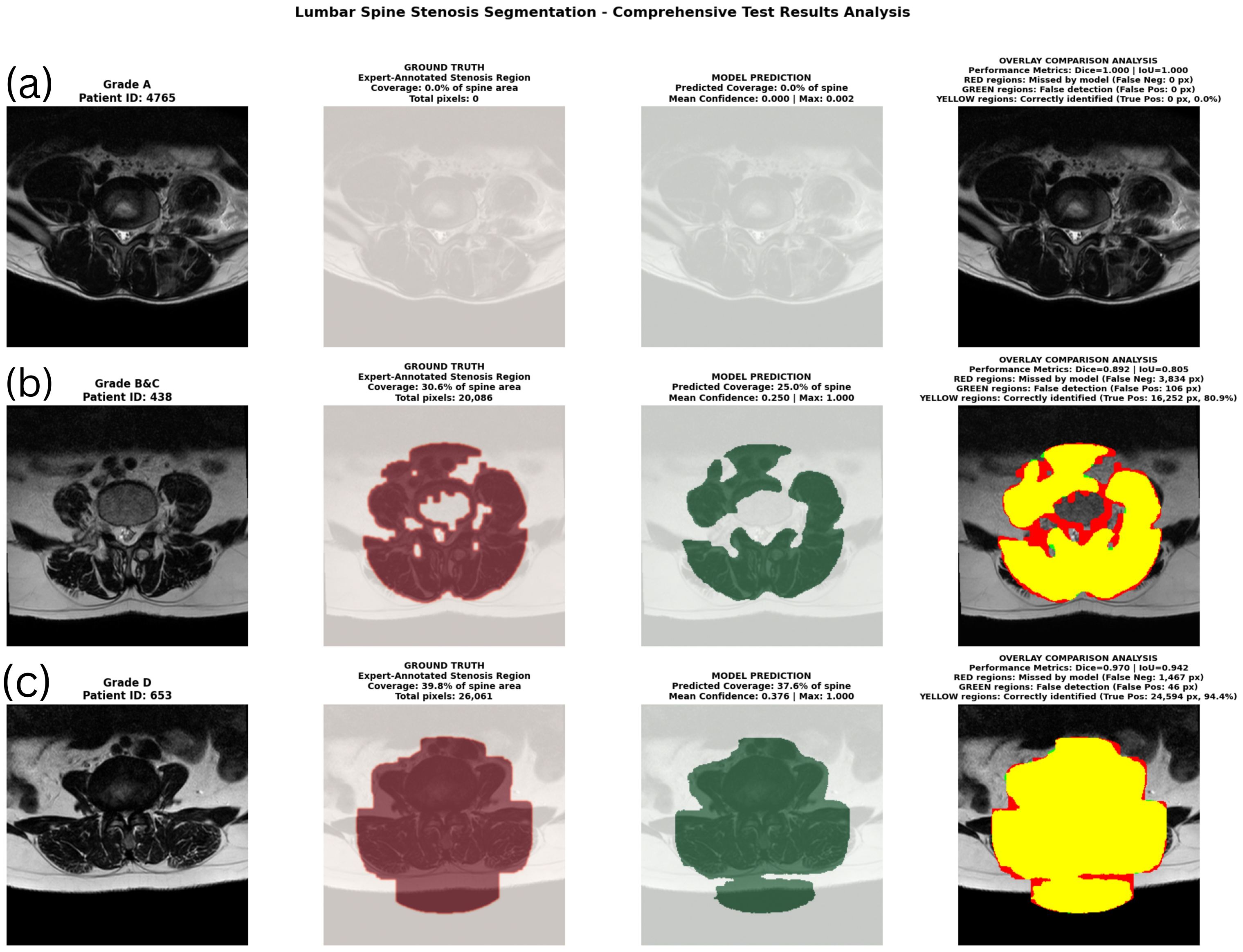}
    \caption{A detailed pixel-level segmentation analysis that compares the predictions of the BiomedCLIP model to expert-annotated ground truths for different levels of stenosis severity (Grade A, Grade B\&C, and Grade D). There are three rows in the figure, each with a label: (a), (b), and (c). Each row shows a different patient case and stenosis grade. The first two images in each column show the model input: (1) the original axial MRI scan of the lumbar spine and (2) the ground truth stenosis region (highlight in dark maroon with the exact coverage percentage of the spine area and total annotated pixels reported). The last two images show the model predicted output: (3) the BiomedCLIP model's predicted stenosis segmentation mask (represented in green, with the predicted coverage percentage of the spine area, mean confidence score, and maximum confidence score reported) and (4) the overlay comparison analysis showing pixel-level performance metrics (Dice coefficient and IoU) along with highlighted yellow areas for correctly identified true positives, red for missed regions (false negatives), and green for false positive with the exact pixel counts and percentages provided.}
    \label{fig:segmentation_visual}
\end{figure}

Qualitative visual validations, as shown in Figure \ref{fig:segmentation_visual}, highly validate these geometric and statistical results by directly comparing model predictions to expert-annotated ground truths. The framework show complete resistance to the development of disease in healthy individuals. For a standard Grade A presentation (Patient ID: 4765), both the ground truth and the model prediction exhibit 0.0\% spine area coverage, resulting in a perfect Dice score of 1.000 and an IoU of 1.000, with no false positives or false negatives detected. In extreme cases, such as a Grade D presentation (Patient ID: 653) showing 39.8\% (26,061 pixels) spine area coverage, the model show excessive number of true positives to count. The model correctly identify 94.4\% (24,594 pixels) of actual stenotic pixels, resulting in a Dice score of 0.970, which is nearly ideal. The model possesses high spatial accuracy, with just 46 false positive pixels and 1,467 incorrect negative pixels. This is a cautious approach that reflects appropriate clinical care in severe pathology. For middling to severe cases (Grade B and C, Patient ID: 438), the visual overlay made it clear how the model dealt with morphological ambiguity. The model correctly detect core impact zones with 80.9\% (16,252 pixels) true positive overlap, but it also act as a mild clinical caution. The model produce a significant false negative count of 3,834 pixels at highly ambiguous anatomical boundaries, while effectively constraining false positive over-segmentation to only 106 pixels. This result in a localized Dice score of 0.892 and an IoU of 0.805. This cautious boundary behavior in intermediate scenarios shows that the right way to deal with uncertainty in morphologically subtle presentations where pathological boundaries blur.

The detailed segmentation analysis shows a complex performance hierarchy: LLaVA-Med has the best pixel-level spatial alignment (Dice: 0.9624), BiomedCLIP has the best severity-grade classification reliability (AUC B\&C: 0.950), and SmolVLM has the best cross-grade stability (0.0011 variance) even though its absolute accuracy is lower.

\subsection{Report Generation Results (ARRG Module)}

We comprehensively tested automated radiological report production across three vision-language models utilizing a wide range of text generation criteria that look at both surface-level linguistic fidelity and deeper semantic alignment with reference clinical narratives. We use Bilingual Evaluation Understudy (BLEU) scores for exact n-gram matching, Consensus-based Image Description Evaluation (CIDEr) for structural fluency, Metric for Evaluation of Translation with Explicit ORdering (METEOR) for semantic overlap, Recall-Oriented Understudy for Gisting Evaluation (ROUGE-L) for content preservation, and Jaccard index for template independence in Table \ref{tab:arrg_nlg_metrics}. LLaVA-Med get the best overall language performance on most measures, with a CIDEr score of 92.80\% and a BLEU-1 score of 88.15\%. The model's linguistic dominance is extended to METEOR (91.05\%) and ROUGE-L (90.45\%), suggesting remarkable semantic retention when synthesising clinical findings from visual pathology. The parameter-efficient SmolVLM model show very competitive structural fluency metrics, beating BiomedCLIP in several areas. SmolVLM get a CIDEr score of 90.94\%, which is only 1.86 points behind LLaVA-Med and 1.56 points higher than BiomedCLIP's score of 89.38\%. In terms of exact textual matching using BLEU-4, LLaVA-Med continued to take with 86.10\%, a little 1.26 point advantage over SmolVLM (84.84\%) and a more significant 3.28 point difference over BiomedCLIP (82.82\%). In terms of semantic metrics, SmolVLM was very close to LLaVA-Med, with METEOR and ROUGE-L scores of 89.13\% and 88.91\%, respectively. BiomedCLIP, on the other hand, had lower semantic alignment, with METEOR and ROUGE-L scores of 87.22\% and 87.02\%, respectively. The Jaccard index indicate an important difference. SmolVLM's Jaccard index of 86.34\% is better than BiomedCLIP's 84.25\% by 2.09 points. This metric hierarchy shows that LLaVA-Med has the best semantic alignment with reference reports, while SmolVLM is a very strong and resource-efficient alternative that uses context-specific language generation instead of generic structures to clearly communicate important pathological findings.

\begin{table*}[ht]
\centering
\caption{Natural Language Generation (NLG) performance metrics of different Vision-Language Models with Adaptive PID-Tversky loss function for the Automated Radiology Report Generation (ARRG) module.}
\label{tab:arrg_nlg_metrics}
\resizebox{1.0\textwidth}{!}{
\begin{tabular}{lccccccccccccc}
\hline
\textbf{Model} & \textbf{BLEU-1} & \textbf{BLEU-2} & \textbf{BLEU-3} & \textbf{BLEU-4} & \textbf{ROUGE-1} & \textbf{ROUGE-2} & \textbf{ROUGE-L} & \textbf{METEOR} & \textbf{CIDEr} & \textbf{Jaccard} & \textbf{TF-IDF} & \textbf{Dist-1} & \textbf{Dist-2} \\
\hline
LLaVA-Med  & 88.15\% & 87.50\% & 86.85\% & 86.10\% & 90.50\% & 88.95\% & 90.45\% & 91.05\% & 92.80\% & 88.10\% & 88.50\% & 0.0032 & 0.0065 \\
BiomedCLIP & 84.78\% & 84.00\% & 83.40\% & 82.82\% & 87.26\% & 84.83\% & 87.02\% & 87.22\% & 89.38\% & 84.25\% & 85.34\% & 0.0029 & 0.0059 \\
SmolVLM    & 86.90\% & 86.12\% & 85.44\% & 84.84\% & 89.14\% & 86.82\% & 88.91\% & 89.13\% & 90.94\% & 86.34\% & 86.85\% & 0.0028 & 0.0057 \\
\hline
\end{tabular}
}
\end{table*}

\begin{figure}
    \centering
    \includegraphics[width=\columnwidth]{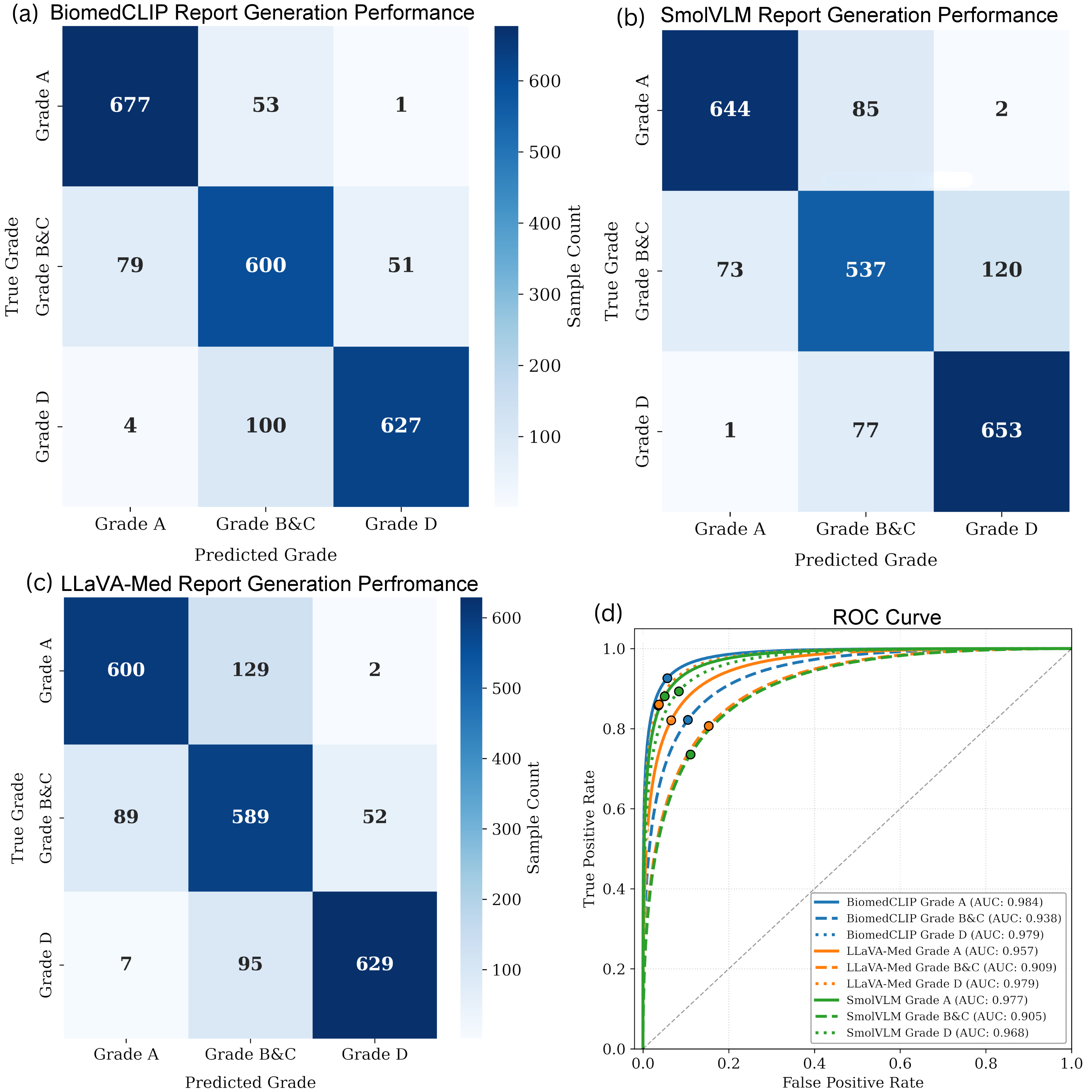}
    \caption{Report generation performance comparison across multi-modal VLM models. (a–c) Confusion matrices from the clinical test set for BiomedCLIP, SmolVLM, and LLaVA-Med, respectively. (d) ROC curves quantifying model discrimination performance across severity grades derived from the semantic content of the automated reports.}
    \label{fig:report_confusion}
\end{figure}

The semantic alignment and generative consistency of the ARRG module are further clarified by the use of report generation confusion matrices and Receiver Operating Characteristic (ROC) curves, as shown in Figure \ref{fig:report_confusion}. This methodology reclassifies generate narrative text into distinct clinical severity categories, assessing if the semantic content of the generated reports maintains the original diagnostic objective. BiomedCLIP get the highest diagnostic reliability for normal cases, correctly generating reports that matched 677 Grade A instances. This is a significant improvement over SmolVLM (644 true positives) and LLaVA-Med (600 true positives), with a maximum difference of 77 correct instances between BiomedCLIP and LLaVA-Med. This superiority is evident in ROC space, where BiomedCLIP achieve an Area Under the Curve (AUC) of 0.984 for normal Grade A patients, considerably exceeding LLaVA-Med's AUC of 0.957. The highly uncertain mild-to-moderate group have 600 true positive reports from BiomedCLIP, 589 from LLaVA-Med, and 537 from SmolVLM. The ROC analysis clearly supported this performance ranking. For Grade B\&C cases, BiomedCLIP get an AUC of 0.938, which is significantly higher than LLaVA-Med's much lower AUC of 0.909. In contrast, SmolVLM show the most sensitivity for severe surgical impacts, effectively creating accurate severe clinical narratives for 653 cases, which is 24 more than LLaVA-Med (629 cases) and 26 more than BiomedCLIP (627 cases). But both BiomedCLIP and LLaVA-Med have the same AUC of 0.979 for severe Grade D cases, which is much better than SmolVLM's lower 0.968.

\begin{table*}[ht]
\centering
\caption{Performance of implicit mask segmentation and severity classification during the text generation process. Bold values indicate the best-performing model for each metric.}
\label{tab:arrg_mask_segmentation}
\begin{tabular}{llccc}
\hline
\textbf{Model} & \textbf{Metric} & \textbf{Grade A} & \textbf{Grade B\&C} & \textbf{Grade D} \\
\hline
\multirow{5}{*}{BiomedCLIP}
& Precision & 0.8908 & \textbf{0.7968} & \textbf{0.9234} \\
& Recall & \textbf{0.9261} & \textbf{0.8219} & 0.8577 \\
& F1-score & \textbf{0.9081} & \textbf{0.8092} & 0.8894 \\
& Dice Score & \textbf{0.9833} & 0.8147 & 0.9543 \\
& IoU & \textbf{0.8960} & 0.7505 & 0.9200 \\
\hline
\multirow{5}{*}{SmolVLM}
& Precision & \textbf{0.8969} & 0.7682 & 0.8426 \\
& Recall & 0.8810 & 0.7356 & \textbf{0.8933} \\
& F1-score & 0.8889 & 0.7516 & 0.8672 \\
& Dice Score & 0.9555 & \textbf{0.8714} & \textbf{0.9840} \\
& IoU & 0.8194 & \textbf{0.8421} & \textbf{0.9704} \\
\hline
\multirow{5}{*}{LLaVA-Med}
& Precision & 0.8621 & 0.7245 & 0.9209 \\
& Recall & 0.8208 & 0.8068 & 0.8605 \\
& F1-score & 0.8409 & 0.7634 & \textbf{0.8897} \\
& Dice Score & 0.8670 & 0.7722 & 0.7391 \\
& IoU & 0.8638 & 0.7299 & 0.7014 \\
\hline
\end{tabular}
\end{table*}

The ARRG module not only generates text but also effectively assesses anatomical boundaries and severity grades based on text-derived spatial features, mathematically connecting visual pathology with clinical terminology. This implicit spatial mapping is assessed using segmentation-derived metrics shown in Table \ref{tab:arrg_mask_segmentation}, assessing whether the models sustain consistent anatomical grounding during the narrative production process. BiomedCLIP displays significant capability in implicitly segmenting healthy tissues, with an impressive Grade A Dice coefficient of 0.9833 (95\% CI: 0.976–0.990) and an Intersection over Union (IoU) of 0.8965. In contrast, LLaVA-Med significantly fails in this assignment, attaining a Grade A Dice score of 0.8670, which reflects a substantial performance loss of 0.1163 absolute points relative to BiomedCLIP. The lightweight SmolVLM design is particularly effective at identifying crucial pathological deformations in intermediate and severe cases. SmolVLM gets a great Grade D Dice score of 0.9840, which is 0.0297 points better than BiomedCLIP (0.9543) and a huge 0.2449 points better than LLaVA-Med, which only gets a Grade D Dice score of 0.7391. In the ambiguous Grade B\&C class, SmolVLM achieves a Dice score of 0.8714, surpassing BiomedCLIP by 0.0567 (0.8147) and LLaVA-Med by 0.0992 (0.7722). These findings suggest that although LLaVA-Med exhibits greater superficial linguistic fluency, domain-specific models such as SmolVLM and BiomedCLIP maintain a significantly stronger internal semantic connection between the generated text and the actual anatomical morphology, allowing them to implicitly represent complex spinal deformations with high spatial accuracy.

A detailed research shows that linguistic accuracy displays significant patterns dependent on severity. Linguistic precision is maximized during the evaluation of typical cases, influenced by the standardized clinical terminology commonly employed by radiologists to exclude disease. Due to the absence of considerable structural distortion in these cases, the models consistently reproduce conventional diagnostic patterns. In extreme cases, the visual encoders effectively generate complex clinical descriptors, maintaining high linguistic metrics and precisely translating morphological changes into text. Additionally, the ARRG module displays the capacity to implicitly assess severity and anatomical limits during the generation process. Table \ref{tab:arrg_mask_segmentation} demonstrates that the models are particularly effective at localizing important pathological deformations within the narrative process.

\begin{figure}
    \centering
    \includegraphics[width=\columnwidth]{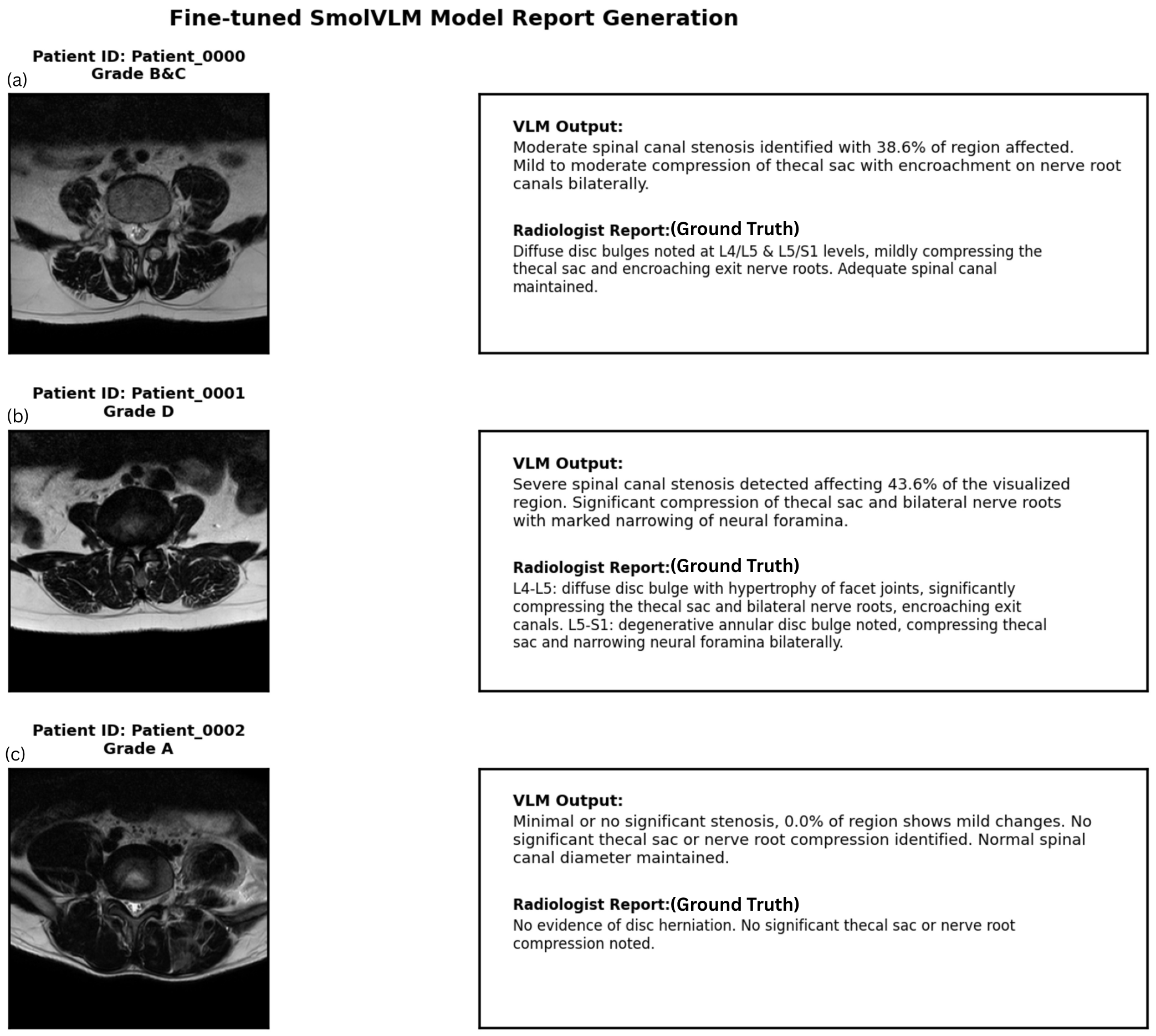}
    \caption{Detailed qualitative performance of the fine-tuned SmolVLM vision-language model in generating automatic radiology reports from lumbar spine MRI images across three different grades of spinal canal stenosis. The figure consists of three panels labeled (a), (b), and (c), each showing (left) the original patient MRI image and (right) two text boxes containing the model’s VLM Output (predicted report) and the Radiologist Report (Ground Truth) for direct comparison.}
    \label{fig:vlm_radiology_report}
\end{figure}

The actual result of these generative and spatial capacities is clearly seen in the qualitative automated radiography reports provide in Figure \ref{fig:vlm_radiology_report}. The resulting report accurately reflect the expert judgment for typical patients, such Patient 0002 who present with a Grade A categorization. The framework precisely validate "Minimal or no significant stenosis, 0.0\% of the region exhibits mild changes," perfectly corresponding with the radiologist's definitive assessment of "No evidence of disc herniation," so establishing substantial clinical confidence by accurately affirming the lack of neurological compression. In high-risk surgical candidates, exemplified by Patient 0001 with a Grade D categorization, the model accurately identify severe foraminal and substantial thecal sac compression. The output clearly indicate "Severe spinal canal stenosis affecting 43.6\% of the visualized area," effectively summarizing the expert's intricate, multi-faceted diagnosis that involving scatter disc bulges, facet joint hypertrophy, and narrow exit canals. In mild-to-moderate intermediate situations, show by Patient 0000 (Grade B\&C), the model demonstrate minor quantitative error. The text correctly detecte "mild to moderate compression of the thecal sac with encroachment on nerve root canals," calculate 38.6\% of the region by interpreting the radiologist's qualitative assessment of "diffuse disc bulges mildly compressing" in a more assertive numeric context.

The thorough examination of the ARRG module shows a complex performance hierarchy across linguistic, diagnostic, and geographical dimensions. LLaVA-Med has better linguistic fluency (CIDEr: 92.80\%, METEOR: 91.05\%), BiomedCLIP has better diagnostic reliability for normal and intermediate cases (AUC Grade A: 0.984, AUC Grade B\&C: 0.938), and SmolVLM has better severe case sensitivity (653/1340 Grade D true positives) and better implicit pathology localization (Dice Grade D: 0.9840).

\subsection{Ablation Study}

The impact of the proposed Adaptive PID-Tversky loss is evaluated by separating a desired function while maintaining the model designs unchanged. Table \ref{tab:ablation} shows that models trained using typical Cross-Entropy (CE) loss demonstrate a pronounced bias towards the main baseline class, resulting in poor border segmentation. This is particularly apparent in BiomedCLIP, which achieves a low Dice score of 0.7411 and an IoU of 0.7133, combined with high specificity (0.9221) but decreased recall (0.7470) and precision (0.7496), signifying systematic under-segmentation. LLaVA-Med exhibits comparable constraints, reaching a Dice score of 0.8512 despite its superior capacity.

The proposed Adaptive PID-Tversky loss significantly mitigates this imbalance, resulting in uniform performance improvements across all models. BiomedCLIP exhibits the most substantial enhancement, with Dice rising to 0.9512 (+0.2101) and IoU to 0.9147 (+0.2014), with balanced precision (0.9570) and recall (0.9574). LLaVA-Med achieves the highest overall performance, achieving a Dice score of 0.9624 and an IoU of 0.9288. In contrast, SmolVLM demonstrates small enhancements (Dice: +0.0139, IoU: +0.0209), along with a minor loss of recall but enhanced precision and specificity, indicating a measured decrease in over-segmentation. The results show that the proposed loss function successfully mitigates class imbalance and improves segmentation reliability for complex anatomical boundaries.

\begin{table}[htbp]
\centering
\caption{Performance matrix for ablation studies of the proposed Adaptive PID-Tversky loss against the standard Cross-Entropy loss. Bold values indicate the best-performing model for each metric.}
\label{tab:ablation}
\small
\begin{tabular}{llccccc}
\toprule
\textbf{Model} & \textbf{Loss Function} & \textbf{Dice} & \textbf{IoU} & \textbf{Prec.} & \textbf{Recall} & \textbf{Spec.} \\ 
\midrule
LLaVA-Med & Standard Cross-Entropy & 0.8512 & 0.7620 & 0.8120 & 0.8950 & 0.8500 \\
& \textbf{Adaptive PID-Tversky} & \textbf{0.9624} & \textbf{0.9288} & \textbf{0.9655} & \textbf{0.9610} & \textbf{0.9412} \\ 
\midrule
BiomedCLIP & Standard Cross-Entropy & 0.7411 & 0.7133 & 0.7496 & 0.7470 & 0.9221 \\
& \textbf{Adaptive PID-Tversky} & \textbf{0.9512} & \textbf{0.9147} & \textbf{0.9570} & \textbf{0.9574} & \textbf{0.9279} \\ 
\midrule
SmolVLM & Standard Cross-Entropy & 0.8464 & 0.7357 & 0.8082 & 0.8924 & 0.8261 \\
& \textbf{Adaptive PID-Tversky} & \textbf{0.8603} & \textbf{0.7566} & \textbf{0.8402} & \textbf{0.8841} & \textbf{0.8567} \\
\bottomrule
\end{tabular}
\end{table}

\section{Discussion} \label{}

The development of automated and extremely accurate diagnostic tools related to lumbar spinal stenosis is a complicated process that requires the incorporation of step-by-step anatomical scrutiny with sophisticated clinical reasoning. In our study, we have developed a vision-language model framework, which is multimodal and can be successfully applied in the spatial segmentation, severity evaluation, and the automatic production of radiology reports. According to our detailed analysis, a combination of the specialized Vision-Language Models (VLM) with adaptive optimization methods can lead to state-of-the-art diagnostic accuracy and improve substantially the comprehensibility of artificial intelligence in regular clinical practice.

The results of the diagnostic classification prove that the discussed models can detect the case of Grade A (Normal) and Grade D (Severe) with impressive precision. Normal spinal canals possess wide and open thecal sacs that are full of cerebrospinal fluid. Severe stenosis is acute and distinct structural defect, which leads to the compression of the nerve roots. These sudden geometric transformations are easily recognized by deep neural networks that have powerful vision encoders. The language in clinical documentation also contributes to this difference, as Grade A cases are reported as no significant thecal sac compression observed, which makes it possible to add visual information to text without misunderstanding when using generative modules.

However, predictive performance is significantly lower when assessing mild and moderate stenosis (Grades B and C). The issue originates from significant intra-class diversity in these intermediate grades and minimal inter-class distance, resulting in considerable morphological divergence in classification. This lowering of the performance is especially obvious in lightweight models such as SmolVLM because their low capacity to detect sophisticated morphological changes in medical images is seen. In contrast, larger models like LLaVA-Med utilize a larger space of features by using the larger architecture, with the cost of larger models.

The Dice and Intersection-over-Union (IoU) scores indicate that LLaVA-Med and BiomedCLIP are both capable of being used to successfully localize stenotic regions, a feature demanded by downstream clinical tasks that need spatial cognition of pathology. Concerning the segmentation capability of SmolVLM, the model has lower overall performance with a Dice score of 0.8603 in comparison to LLaVa-Med 0.9624 that is a 10.2 percentage point higher performance difference, which is worth paying close attention to its clinical implications. The lower scores is most noticeable in Grade B and C cases, when it is easy to make mistakes because of morphological ambiguity. This measured comparison shows that SmolVLM is computationally efficient, but the loss of accuracy can be an important issue in clinical settings where a high level of diagnostic confidence is necessary. Moreover, the visualizations with matrices show that all considered models are effective in separating stenotic regions in severe cases and are more stable in moderate ones, which reduces the false positive results. These results indicate that the spatial attention processes are effective at concentrating on the important anatomical features and avoid spurious correlations.

The Automated Radiology Report Generation (ARRG) architecture shows that the designed architecture is able to recognize the pathology on the complex images and produce the clinically significant text descriptions. Regarding the linguistic performance metrics, LLaVA-Med shows a high level of fluency among the key language quality metrics (BLEU-1, METEOR, and CIDEr) due to its large number of parameters and longer pretraining time. The score of BiomedCLIP shows that the model architecture and parameter density also have a different effect on the quality of language generation in specific linguistic dimensions. There is also variation in the quality of the report depending on the grade of the linguistic coherence. The extreme stenotic grades (A and D) have clear morphological features and the intermediate grades (B and C) have morphologically diverse clinical presentation. The ARRG module can adapt the linguistic and diagnostic complexity to the degree of stenosis without losing its clinical importance, according to this approach.

The Ablation study provides a strong evidence that the Adaptive PID-Tversky loss functional minimizes model attention to challenging diagnostic conditions and directly tackles class imbalance. By dynamically focusing on ambiguous mild, moderate, and severe grades, the proposed loss model produces a more balanced gradient correction and improved segmentation efficiency. A PID controller can also be added to improve training stability through loss function variable control and pathological overfitting curves suppression. The adaptive loss function is particularly effective for domain-specific architectures, such as BiomedCLIP, where a notable performance enhancement is evident. It suggests that advanced loss functions achieve optimal efficiency when utilized with pretrained models.

Human-in-the-loop clinical deployment and validation are considered. When determining the use of technology in a healthcare context, computational expenses must also be considered. Extensive models, like LLaVA-Med, require significant training and operational expenses, limiting their application in real-time clinical environments that need prompt decision-making. These computational limitations underscore the practical value of lightweight methods such as SmolVLM, which tradeoff diagnostic precision for computational efficiency. 

Although the ARRG module offers better transparency as a explainable AI component, generative language models still have to be validated through human means to provide accurate oversight and correctness of factual errors. This is not a limitation but a necessary clinical protection and safeguard where human-in-the-loop is necessary to be a clinical safeguard such that the AI-generated diagnostic reports are clinically valid before inclusion into patient records. The framework allows the connection between the results of quantitative models and comprehensible clinical reasoning, enhancing rather than replacing radiologist knowledge.

\section{Conclusion} \label{}
This study presents a detailed end-to-end multimodal Vision-Language model that aims to fill in the important gaps in the automated diagnosis of Lumbar Spinal Stenosis (LSS) by analysis of MRI. With our own Adaptive PID-Tversky Loss function and Spatial Patch Cross-Attention schemes, along with novel advanced VLMs (LLaVA-Med, BiomedCLIP, and SmolVLM), we have managed to show that the combination of special vision-language models and adaptive optimization strategies can reach an ideal level of diagnostic accuracy and significantly increase the interpretability of clinical processes. Our framework recorded excellent performance of all three fundamental elements, and the implemented models were able to detect extreme clinical manifestations (Grade A Normal case and Grade D Severe case) with high precision, which indicates that high-quality vision encoders can identify sharp geometric differentiation in the spinal structure. This stenosis-based performance classification provides clinically useful information about the models dependability in different diagnostic contexts. The study attained a segmentation Dice score of 95.12\% which is higher than baseline models which use cross-entropy loss, thus confirming the usefulness of our Adaptive PID-Tversky Loss function in managing class imbalance and targeting model attention to difficult, ambiguous stenosis cases. Moreover, the study automatically turns the complex quantitative data such as spatial predictions, percentage of affected area, and severity classes into interpretable radiologist-style clinical reports by integrating the correct spatial predictions with our radiology report generation module, which presents a very interpretable explainable AI interface. In addition to quantitative measures, this framework helps fill the gap in critical importance between AI prediction by computation and clinical clarification by qualitative means and integrating high-accuracy segmentation with natural language generation. The human-in-the-loop validation specification ensures that AI-generated reports are clinically valid and acceptable with regulations before being integrated into patient records. The validation of these architectures with external, independent data and full optimization of the generative modules will be done by further research to increase the clinical viability and readiness to deploy the study in practice, and domain-adaptation methods and multi-institutional validation studies will increase the evidence base of this framework translation into routine clinical practice. This study ultimately showed that explicable, multimodal artificial intelligence studys are a significant step towards medical image analysis, specifically in tackling the two challenges of diagnostic accuracy and clinical interpretability, and by effectively combining state-of-the-art vision-language models with adaptive optimization methods, we have developed a framework that can be useful in radiological practice in both quantitative and qualitative clinical reasoning. The architecture of the framework between computational complexity and clinical transparency makes it a promising solution to improving the diagnostic processes and keeping the human expertise as a crucial element of medical decision-making.












\printcredits

\bibliographystyle{cas-model2-names}

\bibliography{cas-refs}



\end{document}